\documentclass[letterpaper]{article} 
\usepackage{aaai2026}  
\usepackage{times}  
\usepackage{helvet}  
\usepackage{courier}  
\usepackage[hyphens]{url}  
\usepackage{graphicx} 
\usepackage{natbib}  
\usepackage{caption} 
\usepackage{algorithm}

\usepackage{newfloat}
\usepackage{listings}
\DeclareCaptionStyle{ruled}{labelfont=normalfont,labelsep=colon,strut=off} 
\floatstyle{ruled}
\newfloat{listing}{tb}{lst}{}
\floatname{listing}{Listing}
\usepackage{xspace}
\usepackage{pifont}
\usepackage{amssymb}
\usepackage{enumerate}
\usepackage{subcaption}
\usepackage{multirow}
\usepackage{totcount}
\usepackage{xcolor,colortbl}
\usepackage{algpseudocode}
\usepackage{enumitem}
\usepackage{booktabs}
\usepackage{amsmath}

\newcommand{\specialcell}[2][c]{%
  \begin{tabular}[#1]{@{}l@{}}#2\end{tabular}}

\title{Towards Scalable Web Accessibility Audit with MLLMs as Copilots}
\author {
    Ming Gu\textsuperscript{\rm 1,\rm 2},
    Ziwei Wang\textsuperscript{\rm 1,\rm 2},
    Sicen Lai\textsuperscript{\rm 1,\rm 3},
    Zirui Gao\textsuperscript{\rm 1,\rm 2},
    Sheng Zhou\textsuperscript{\rm 1,\rm 3}\thanks{Corresponding Author},
    Jiajun Bu\textsuperscript{\rm 1,\rm 2}
}
\affiliations {
    \textsuperscript{\rm 1}Zhejiang Key Laboratory of Accessible Perception and Intelligent Systems, Zhejiang University\\
    \textsuperscript{\rm 2}College of Computer Science and Technology, Zhejiang University\\
    \textsuperscript{\rm 3}School of Software Technology, Zhejiang University\\
    \{gmwork, wangziwei98, laisicen, gaozirui.zju, zhousheng\_zju, bjj\}@zju.edu.cn
}

\usepackage{bibentry}

\begin{document}

\maketitle
\begin{abstract}
Ensuring web accessibility is crucial for advancing social welfare, justice, and equality in digital spaces, yet the vast majority of website user interfaces remain non-compliant, due in part to the resource-intensive and unscalable nature of current auditing practices. While WCAG-EM offers a structured methodology for site-wise conformance evaluation, it involves great human efforts and lacks practical support for execution at scale. In this work, we present an auditing framework, AAA, which operationalizes WCAG-EM through a human-AI partnership model. AAA is anchored by two key innovations: GRASP, a graph-based multimodal sampling method that ensures representative page coverage via learned embeddings of visual, textual, and relational cues; and MaC, a multimodal large language model-based copilot that supports auditors through cross-modal reasoning and intelligent assistance in high-effort tasks. Together, these components enable scalable, end-to-end web accessibility auditing, empowering human auditors with AI-enhanced assistance for real-world impact.
We further contribute four novel datasets designed for benchmarking core stages of the audit pipeline.
Extensive experiments demonstrate the effectiveness of our methods, providing  insights that small-scale language models can serve as capable experts when fine-tuned.
\end{abstract}

\newcommand{\figureautorefname}{Figure}
\newcommand{\tableautorefname}{Table}
\newcommand{\framework}{AAA\xspace}
\newcommand{\grasp}{GRASP\xspace}
\newcommand{\mac}{MaC\xspace}
\newcommand{\data}{AWA\xspace}

\begin{links}
    \link{Code \& Datasets}{https://github.com/eaglelab-zju/AAA}
    \link{Standard version}{https://openreview.net/forum?id=kz2hcsWcGu}
\end{links}

\section{Introduction}

\textit{Web accessibility} is a foundational principle in the pursuit of an inclusive digital environment, ensuring that all users including those with disabilities can perceive, navigate, and interact with online content~\cite{w3c,sys2}.
Despite the widespread adoption of standards such as the Web Content Accessibility Guidelines (WCAG)~\cite{wcag}, and substantial efforts devoted to web accessibility evaluation, the state of web accessibility remains alarmingly poor.
A recent study reported that 94.8\% of homepages across one million websites contained accessibility violations~\cite{webaim}.
Emerging research suggests that this stagnation stems not from a lack of education or tooling, but from the intrinsic complexity of web accessibility as a resource management problem~\cite{duo,beyond}.
This means that\textit{ the time-consuming and labor-intensive nature of Web Accessibility Audits (WAA) increasingly misaligned with the growing scale and maintenance cost of modern websites}~\cite{large}.

To address WAA, the World Wide Web Consortium (W3C) introduced the Website Accessibility Conformance Evaluation Methodology (WCAG-EM)~\cite{em}, a five-step protocol designed to standardize evaluation procedures.
However, it lacks a corresponding technical framework that supports scalable execution in practice.
In this context, \textbf{scalability} refers to two critical capabilities: (1) accelerating audit processes via automation, and (2) minimizing unavoidable manual effort through intelligent human–AI collaboration.
Yet, most existing tools operate only at the page or element level, \textit{covering only fragments of the WCAG-EM pipeline}~\cite{correcting}. 
This narrow scope hinders scalability with bottlenecks in both time and labor. 

To overcome the limitations, we propose a comprehensive framework anchored in three pillars: Automation, AI, and Auditor (\framework).
\textit{\framework operationalizes five procedures aligned with WCAG-EM's five steps}, including web crawling, automated checks, page sampling, manual evaluation, and reporting/remediation, with the goal of enabling scalability across the full audit lifecycle.
Despite advances in automating tasks  such as crawling and hard-coded checks,\textit{ two fundamental challenges remain}.
\textbf{First}, existing page sampling methods fail to satisfy WCAG-EM's representativeness requirements. 
Recent clustering-based approaches rely primarily on textual similarity~\cite{ps}, overlooking the rich multimodal semantics of web pages including visual layout, textual content and hyperlink relationships, which are essential for capturing diversity and representativeness.
\textbf{Second}, intelligent assistance for manual auditing tasks remains underexplored. 
Given that no single tool can fully determine whether a website meets accessibility standards~\cite{auditor}, WAA inevitably requires human evaluation. 
However, current methods offer minimal assistance in high-effort tasks such as identifying accessibility-critical components, which often demand sophisticated multimodal reasoning, making them particularly burdensome for human auditors to collect.

To tackle these challenges, we first introduce \underline{G}raph-based \underline{R}epresentative P\underline{A}ge Clustering for \underline{S}am\underline{P}ling (GRASP), a novel multimodal approach that generates WCAG-EM-compliant representative page subsets.
GRASP defines representativeness across three complementary dimensions: textual semantic, visual layout, and linkage relationships, and employs graph neural networks (GNNs) to learn a unified embedding space for representative clustering. 
A dedicated structure learning module further improves sampling quality by mutually enhancing representativeness and clustering.
In parallel, we explore the emerging potential of multimodal large language models (MLLMs) in accessibility workflows~\cite{sr}. 
We present \underline{M}LLMs \underline{a}s \underline{C}opilot Assistant, Auditor and Consultant (MaC), a holistic AI companion designed to support multiple stages of WAA.
By enabling cross-modal reasoning, MaC assists in identifying audit-critical elements and pages, thereby accelerating both sampling and manual evaluation. 
Furthermore, it broadens audit coverage by facilitating the evaluation of underrepresented accessibility issues, particularly those affecting less commonly addressed disabilities like cognitive impairment.
To support future research, we also release four new datasets tailored for distinct stages of the WAA pipeline.
These datasets address the current lack of accessibility-specific benchmarks.
Our contributions are summarized as: 
\begin{itemize}
    \item \textbf{Scalable WAA Framework}: We propose a full-lifecycle audit framework AAA aligned with WCAG-EM, advancing scalability across web accessibility audit lifecycle.
    \item \textbf{Multimodal Sampling Method}: We introduce GRASP, a novel graph-based multimodal page sampling technique satisfying WCAG-EM representativeness criteria.
    \item \textbf{MLLM as Copilot Strategy}: We introduce MAC, a versatile MLLM-powered strategy augmenting multiple labor-intensive procedures via multimodal reasoning.
    \item \textbf{Benchmark Datasets}: We release four new datasets tailored to different stages of the AAA pipeline, facilitating comprehensive evaluation and comparison. 
    \item \textbf{Empirical Insights}: Through extensive experiments, we demonstrate the effectiveness of our methods and uncover the potential of small MLLMs as domain experts.
\end{itemize}

\section{Related Work}

\textbf{Web Accessibility Audit (WAA)}.
With growing demand for an inclusive web, web accessibility auditing has become essential.
Traditional tools like WAVE~\cite{wave} and Axe~\cite{axe} rely on hard-coded checks that detect syntactic issues (e.g., missing alt text, low contrast), but often missing contextual and semantic aspects~\cite{turning,sr}.
Recent advances leverage large language models (LLMs) to provide intelligent evaluation and repair suggestions, aiming to reduce manual effort~\cite{correcting,case}.
Nevertheless, accessibility gaps persist: a 2025 audit of one million websites revealed WCAG violations on 94.8\% of home pages~\cite{webaim}. 
Studies suggest this stems less from tool limitations and more from resource constraints~\cite{duo}.
Auditing remains time- and labor-intensive, especially as website scale increases nowadays~\cite{large}. 
\textit{Existing approaches predominantly focus on individual elements or pages, lacking a framework to \textbf{address labor and resource challenges across the full site-wise audit lifecycle}.}

\textbf{Page Sampling for WAA}.
Full-site evaluation is infeasible for large sites, making page sampling essential for producing representative results. 
WCAG-EM outlines two dimensions: (1) \textit{the individual level}, which targets pages with critical accessibility relevance (e.g., essential functionality, accessibility statements, or home-linked common pages), and (2) \textit{the collective level}, which ensures diversity and representativeness across the site.
However, existing methods often address narrow aspects, such as URL patterns~\cite{url}, Web Accessibility Quantitative Metric~\cite{waqm}, or structure-based active learning~\cite{multi}, falling short of multi-level requirements.
A recent method, Web Structure Derived Clustering (SDC)~\cite{ps}, attempts to align with collective-level sampling by clustering.
Yet, \textit{it excludes individual-level sampling, potentially omitting accessibility-critical pages, and relies solely on shallow statistical textual features, lacking semantic depth and multimodal integration}.

\textbf{LLM Applications in WAA}.
Several studies have explored the use of LLMs for web accessibility evaluation~\cite{correcting,case,webaudit}, demonstrating impressive automation in addressing element- and page-level issues.
However, existing approaches mainly focus on text semantic alignment~\cite{screenaudit}, such as text or code generation, and \textit{rarely involve more modularities and complex reasoning about accessibility knowledge which multimodal LLMs (MLLMs) are capable of}.
\textbf{Moreover}, most research on LLMs is confined to evaluation and remediation tasks, \textit{overlooking the broader applicability across the entire WAA lifecycle addressing numerous labor-intensive steps}.

\section{AAA: Scalable WAA Framework}

\begin{figure*}[t]
    \centering
    \includegraphics[width=.8\textwidth]{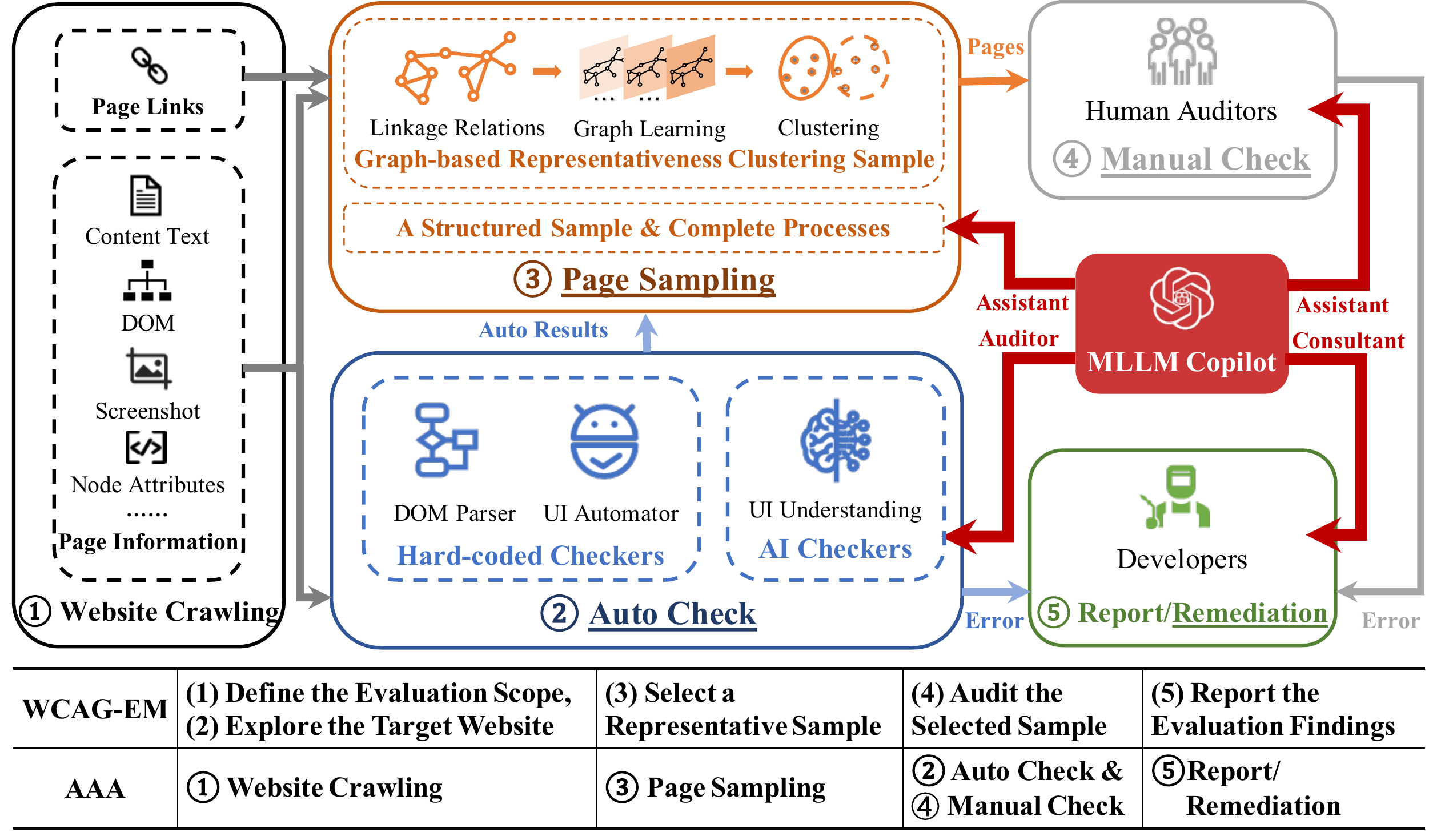}
    \caption{Overview of \framework.
    }
    \label{fig:aaa}
\end{figure*}

\subsection{Pipeline of the Proposed Framework \framework}

We propose a scalable WAA framework for large or multiple sites centered on \underline{A}utomation, \underline{A}I, and \underline{A}uditor (AAA).
Here, \textit{AI} denotes \textbf{\textit{artificial intelligence}} technologies such as computer vision and natural language processing, which can understand abstract concepts beyond rule-based programmatic \textit{\textbf{automation}}.
Moreover, since no single tool can independently determine whether a website meets accessibility standards~\cite{compliance}, and given the reliability and application challenges associated with LLMs, such as hallucinations~\cite{challenges}, knowledgeable evaluation by human \textit{\textbf{auditors}} remains essential.
Inspired by the five-step guidance of WCAG-EM~\cite{em}, \framework is designed for \textit{scalable auditing of large websites or multi-sites, where exhaustive assessment of all content is impractical}.
We reinterpret WCAG-EM from a technical perspective and organize it into five structured procedures.
An overview of \framework along with a comparison to WCAG-EM is in \figureautorefname~\ref{fig:aaa}.

\textbf{Website Crawling}.
For large-scale or multi-site evaluations, manually defining the evaluation scope, as required by WCAG-EM, is impractical.
To address this, automated website crawling is introduced to systematically explore and extract site structures and content at scale.

\textbf{Auto Check}.
Automated checks in \framework are performed using two types of checkers:
\ding{182} \textit{Hard-coded checkers}, which include tools based on static DOM parsing (e.g., Axe~\cite{axe}) and dynamic UI testing frameworks (e.g., Selenium~\cite{sel}).
These tools provide deterministic accuracy but are limited in assessing semantic-level issues.
\ding{183} \textit{AI-powered checkers}, which leverage intelligent technologies to perform visual and textual semantic analysis, enabling the detection of accessibility violations that require deeper contextual understanding.
While accuracy trade-offs are introduced due to the black-box nature, significant potential are offered for evaluating complex issues.

\textbf{Page Sampling}.
WCAG-EM prescribes three steps to construct a representative sample: (i) Include a structured sample, (ii) Include a randomly selected sample, (iii) Include complete processes.
Except for random sampling, these steps require sophisticated semantic understanding of web pages at multiple levels. 
To alleviate the labor burden, we introduce:
\ding{182} A novel deep-learning method integrating visual, textual, and relational information to optimize the \textit{random selection of representative diverse pages}.
\ding{183} The use of MLLMs to \textit{recognize structured samples and complete processes}, leveraging their strong multi-modal semantic understanding capabilities~\cite{understand}.

\textbf{Manual Check}.
While manual evaluation follows WCAG 2.2~\cite{wcag,sc} success criteria, its scalability and coverage remains a challenge.
Two key optimizations are introduced by MLLM-powered assistance:
\ding{182} \textit{Pre-extraction of accessibility-critical items}: Some critical accessibility issues occur infrequently across a website, potentially reducing audit coverage in a sampling-based pipeline. 
MLLMs assist in identifying these elements in advance to ensure a faster and comprehensive evaluation.
\ding{183} \textit{Automation of evaluating underrepresented accessibility issues}: Leveraging the powerful multi-modal reasoning capabilities of MLLMs~\cite{reason}, we can automate the evaluation of accessibility issues that typically receive less attention.

\textbf{Report/Remediation}.
The remediation process is integrated into the reporting step for rich and actionable feedback with automated remediation potentials.
First, reports should contain detailed violation descriptions and developer-friendly repair suggestions.
Second, recent advances in AI-driven code generation provide promising avenues for automated accessibility fixes.

\subsection{Challenges in Implementing \framework}

\textbf{Representative Page Sampling for Scalable Audits}.
\textbf{\textit{First}}, existing approaches to page sampling in WAA are predominantly based on statistical analysis of DOM text~\cite{ps}, which lack a deeper understanding of multimodal semantics like \textit{visual styles and layouts}, \textit{textual topics, and functional diversity}.
These dimensions are essential for identifying representative pages, as emphasized in WCAG-EM Step 2.c ``\textit{Identify the variety of web page types}".
\textbf{\textit{Second}}, with the advancement of MLLMs in comprehending complex semantics across modalities, it is now feasible to automatically identify many key web page types previously reliant on manual inspection, like common web pages and essential functionality in WCAG-EM Step 2.a ``\textit{Identify common web pages}" and Step 2.b ``\textit{Identify essential functionality of the website}".
MLLMs enables the construction of page samples that are not only statistically representative but also semantically aligned with accessibility requirements, largely saving labor and resource.
 
\textbf{Comprehensive MLLMs Integration in WAA}.
\textbf{\textit{First}}, most existing works on LLMs-powered WAA only explore checks that have been well covered by existing automated tools and limited within textual information~\cite{screenaudit,webaudit}, leaving out evaluation that relies on complex modalities or semantics, which is the unique advantage of MLLMs.
\textbf{\textit{Second}}, the existing work on the application of MLLMs in WAA is mostly limited to evaluation or redediation~\cite{codeg}, and has not fully explored its potential in the entire lifecycle.

\textbf{Datasets for WAA Benchamarking}.
Standardized evaluation datasets are still lacking, which include not only cases of accessibility issues that are not detectable by automated tools, but also an evaluation of the application of MLLMs in other steps of the WAA lifecycle.

\section{GRASP: Graph-based Page Sampling}

\begin{table}
    \centering
    \small
    \setlength{\tabcolsep}{1mm}
        \begin{tabular}{lllc}
        \toprule[1pt]
             \textbf{Sub-step}& \textbf{Type} & \textbf{Description} & \textbf{Methods}\\
        \midrule[0.8pt]
        \multirow{5}{*}{\textbf{\underline{3.a}}\quad\rotatebox[origin=c]{90}{Structured Sample}} &\specialcell{Common\\Web Pages\\and States}  & \specialcell{Linked Directly from\\the Main Entry\\Point (Home Page)} & \multirow{10}{*}{\textbf{\mac}} \\
             \cmidrule{2-3}
             & \specialcell{Relevant\\Web Page\\and States} &  \specialcell{Relevant for People\\with Disabilities and\\the Accessibility of\\the Website} & \\
             \cmidrule{2-3}
             & \multirow{3}{*}{\specialcell{Additional\\Web Pages\\and States}} & Essential Functionality & \\
             \cmidrule{3-3}
             &  & \specialcell{Web Technologies\\Relied Upon} & \\
             \cmidrule{3-4}
             &  & \specialcell{Variety of Web\\Page Types}& \textbf{\grasp}\\
             \cmidrule{1-4}
             \multicolumn{2}{l}{\textbf{\underline{3.b}}\quad\specialcell{Randomly\\Selected Sample}}& \specialcell{Number: 10\% of\\the 3.a sample} & —\\
             \hline
             \multicolumn{2}{l}{\textbf{\underline{3.c}}\quad\specialcell{Complete\\ Processes}}& \specialcell{Pages belonging to\\a series presenting\\a complete process} & \textbf{\mac} \\
        \bottomrule[1pt]
        \end{tabular}
    \caption{Three Sub-steps (3.a, 3.b, 3.c) of WCAG-EM Step 3 Achieved by Our Proposed Methods \mac and \grasp.}
    \label{tab:sampling}
\end{table}
\begin{figure*}[t]
    \centering
    \includegraphics[width=.75\textwidth]{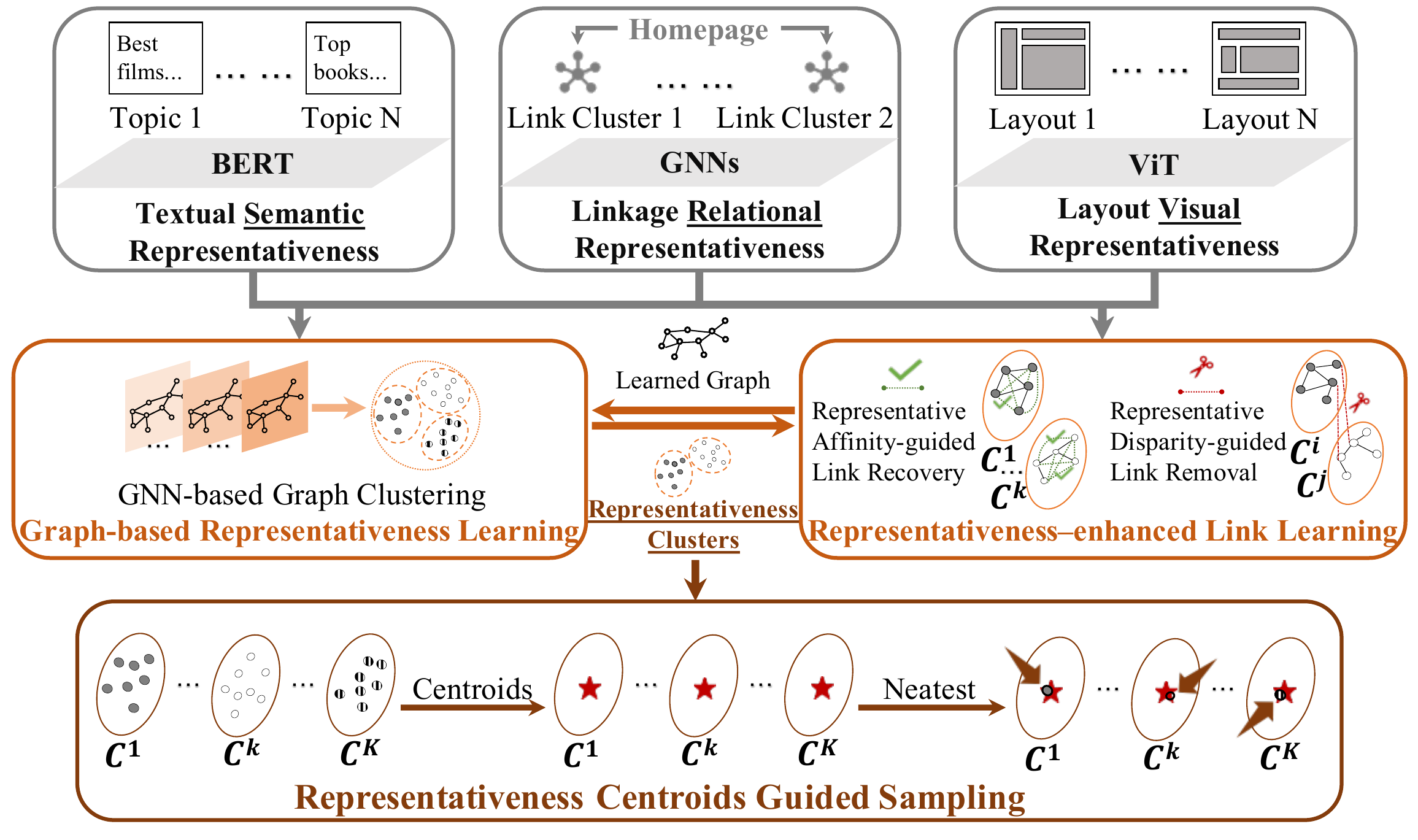}
    \caption{Overview of \grasp. 
    }
    \label{fig:grasp}
\end{figure*}

WCAG-EM Step 3 (S3, \textit{"Select a Representative Sample"})  has three sub-steps as shown in \tableautorefname~\ref{tab:sampling}, in which the randomly selected sample in 3.b can be trivially realized as a random sample with certain tools~\cite{wcag}
.
Therefore, we focus on the other two sub-steps, which are achieved by a two-fold method.
From the \textit{\textbf{individual perspective}}, when the selection of each page is based on the individual characteristic of itself, MLLMs are used as an alternative to the labor-intensive human recognition of (1) common and relevant web pages and states, (2) two kinds of additional web pages and states and (3) all pages of complete processes.
From the \textit{\textbf{collective perspective}}, as tasks not solvable by scale is a well-known challenge faced by MLLMs~\cite{challenges}, which calls for other UI understanding models to deal with this task where a whole picture of different pages across the website is needed to be captured for sampling representative pages among them.
Individual sampling will be introduced in next section, and for collectivitive sampling, we propose \underline{G}raph-based \underline{R}epresentative p\underline{A}ge clu\underline{S}tering for sam\underline{P}ling (\grasp). 
\figureautorefname~\ref{fig:grasp} presents the overview of \grasp.

\subsection{Triple Representativeness for Page Variety}
S3 defines the variety of web page types as \textit{varying styles, layouts, structures, and functionality with varying support for accessibility}~\cite{em}.
These variety representativeness calls for understanding of multimodal semantics, like visual and textual, which are not fully covered in existing researches of statistical analysis.
\grasp addresses this issue by taking into account three types of page representativeness from the perspectives of text, layouts and linkages.

\textbf{Textual Semantic Representativeness}.
\textit{Traditional approaches based on token frequency or lexical analysis fall short, as they tend to reflect only surface-level textual features, neglecting deeper semantic structures and functional intentions.}
To address this, we leverage BERT~\cite{bert}, a contextualized language model that captures the nuanced meaning of words and phrases within their broader linguistic and structural context. 
For instance, it distinguishes between the use of ``submit" in a login form versus in a feedback module by attending to nearby content and structural patterns.
This capability makes BERT particularly well-suited for extracting semantically representativeness.

\textbf{Layout Visual Representativeness}.
With the advancement of web technologies, \textit{the textual structure of the DOM has become increasingly inadequate for reflecting the rendered visual layout of modern web pages—especially in dynamically generated single-page applications (SPAs).}
To overcome this challenge, we employ Vision Transformers (ViT)~\cite{vit} to learn visual representations directly from page screenshots.
This approach enables robust extraction of layout-level representativeness, capturing the spatial and visual organization of web content beyond what is available through static DOM analysis.

\textbf{Linkage Relational Representativeness}.
Moreover, websites inherently contain rich hyperlink structures that are often overlooked in existing approaches. 
These linkages naturally form a graph structure that encodes functional relatedness and semantic proximity across multiple pages.
\textit{For instance, pages belonging to the same functional module frequently exhibit clustering behavior within the hyperlink network, while pages with similar layouts tend to share common linking patterns or structural relationships.}
To model this, we adopt Graph Neural Networks (GNNs), which are well-suited for learning structured data~\cite{sage}.
GNNs support the integration of node attributes with topological context, making them ideal for capturing and fusing this third modality of representativeness.

\subsection{Representativeness-enhanced Page Sampling}

\textbf{GNN-based Graph Representativeness Clustering}.
Recent clustering-based page sampling approaches~\cite{ps} leverages shallow statistical representations derived from textual content, 
which fail to capture the deeper semantic nuances of text, and more critically, neglect both the visual structure and inter-page relational context.
To address this, we introduce a GNN-based graph clustering approach that explicitly integrates multiple modalities into the clustering process.
Our method consists of two key stages.
(1) Modality-specific Representation Learning,
(2) Semantic Fusion via GNN Message Passing.
The first stage is:
\begin{equation}
    \mathbf{H}_t = \text{BERT}(\textrm{text}_\texttt{{DOM}}),\ \mathbf{H}_v = \text{ViT}(\textrm{image}_\texttt{{screen}}),
\end{equation}
\begin{equation}
    \mathbf{X}=\mathbf{H}_t||\mathbf{H}_v, \mathbf{H}_g = \text{GNN}(\mathbf{X}, \mathbf{A}), \mathbf{C} = \mathcal{C}(\mathbf{H}_g),
\end{equation}
where $||$ is concatenation, and $\mathbf{A}$ is the adjacent matrix of the hyperlinks.
$\mathcal{C}$ is a clustering method like k-means, and $\mathbf{C}$ is the clustering assignments of web pages.
\textit{This allows us to learn fused embeddings that encapsulate the combined representativeness across all modalities.}

\textbf{Representativeness-enhanced Graph Learning}.
While hyperlink structures offer a natural foundation for graph-based modeling of websites, they often suffer from noise and sparsity.
\textit{\textbf{First}}, pages with dissimilar semantics may still be interconnected due to structural conventions, such as universally linked footer pages, which do not necessarily indicate semantic relatedness. 
\textbf{\textit{Second}}, pages that are semantically similar may not be directly connected, especially in hierarchical site architectures where linkage primarily reflects parent-child relationships.
Sibling pages with shared intent may remain unlinked by multi-hops in the graph.

We propose a representativeness-enhanced graph learning approach that refines the raw hyperlink structure using \textit{representativeness affinity and disparity} inspired by graph structure learning~\cite{hole,s1}.
Specifically, we leverage the representativeness clustering results derived from textual and visual representation to guide the reconstruction of the linkage graph.
This refinement involves both the removal of noisy links and the recovery of semantically meaningful missing ones.
Formally, we define the hyperlink edge removal and recovery sets as:
\begin{equation}
    \mathcal{E}_\text{rm} = \mathcal{S}_\text{sim}(\mathbf{C}, \mathbf{H}_g, \gamma), \mathcal{E}_\text{rc} = \mathcal{S}_\text{dis}(\mathbf{C}, \mathbf{H}_g, \beta),
\end{equation}
\begin{equation}
    \mathcal{E}_\text{new} = (\mathcal{E}_\mathbf{A}\cap\mathcal{E}_\text{rc})\setminus\mathcal{E}_\text{rm},
\end{equation}
where $\gamma$ and $\beta$ control the thresholds for the least and most semantically similar node pairs, respectively, as determined by similarity functions $\mathcal{S}_\text{sim}$ and $\mathcal{S}_\text{dis}$ operating over cluster assignments $\mathbf{C}$ and node embeddings $\mathbf{H}_g$.
\textit{The set $\mathcal{E}_\text{rm}$ identifies representativeness-disparity guided redundant or misleading links to be pruned from the original edge set $\mathcal{E}_\mathbf{A}$, while $\mathcal{E}_\text{rc}$ identifies representativeness-affinity guided strong candidate connections. }
The refined set $\mathcal{E}_\text{new}$ is obtained by adding promising edges and removing low-quality ones.

\textbf{Representative Centroids-based Sample Selection}.
We perform representative page sampling by selecting exemplar nodes from each cluster.
For each cluster $c_i$ of $\mathbf{C}$, we first compute its centroid $\mu_i$ and then select the representative node $v_i$ closest to the centroid.
The final sampled page set is obtained by aggregating all selected nodes across clusters:
\begin{equation} v^*_i = \arg\min_{v \in c_i} \|\mathbf{H}_g(v) - \mu_i\|_2, \mathcal{P}_\text{sample} = \bigcup_{i} v^*_i.\end{equation}
\textit{This strategy ensures that the sampled pages are deemed the most representative of the semantic, visual, and relational characteristics captured by their cluster.}

\section{Proposed MLLMs Strategies and Datasets}

\begin{table*}[t]
    \centering
    \small
    \setlength{\tabcolsep}{1mm}
    \begin{tabular}{llllll}
    \toprule[1pt]
        \textbf{Dataset} & \textbf{Task} &  \textbf{Type} & \textbf{Input} & \textbf{Output} & \textbf{Method} \\
        \midrule[0.8pt]
        TPS & \specialcell{Page\\Sampling} &\specialcell{ Multimodal\\sampling} & \specialcell{Crawled pages’ DOMs,\\screenshots,\\auto-check results,\\linkage graph, and\\sample size N} & N representative pages & GRASP\\
        \hline
        APR & \specialcell{Accessibility-\\Relevant\\Page Recognition} & Classification & Page screenshots & \specialcell{Prediction of 5\\WCAG-EM-defined\\categories (common,\\relevant, essential,\\technology-dependent,\\or none)} & MaC\\
        \hline
        CCT & \specialcell{Cognitive\\CAPTCHA\\Tests} & \specialcell{Recognition, reasoning\\and classification} & CAPTCHA screenshots & \specialcell{(1)Binary WCAG compliance,\\(2)violation reasons,\\(3)type (17 classes)} & MaC\\
        \hline
        CPE & \specialcell{Complete\\Process\\Extraction} & Multi-label classification & Page screenshots & \specialcell{Presence of 5\\WCAG-EM-defined\\processes (search, filter,\\form, CAPTCHA, contact) }& MaC \\
        \bottomrule[1pt]
    \end{tabular}
    \caption{Task definitions and their relationships with datasets and methods.}
    \label{tab:tasks}
\end{table*}
\subsection{MaC: MLLMs as Various Copilots}

Current applications of LLMs in accessibility face two limitations.
(1) \textit{\textbf{Limited Exploration of Evaluation.}} 
Most existing applications focus on a narrow range of rules, often overlapping with those addressed by traditional tools (e.g., missing alt text)~\cite{webaudit}, while complex tasks involving cognitive accessibility remain largely unaddressed. We argue that the true potential of MLLMs lies in enabling fairer and more comprehensive audits by filling evaluation gaps for \textit{\textbf{underrepresented disabilities} through expert-informed multimodal reasoning}.
(2) \textit{\textbf{Restricted Scope of Applications.}} 
The application of MLLMs in web accessibility remains underexplored beyond evaluation and remediation, particularly in resource-intensive stages like page sampling and manual auditing.
We argue that integrating MLLMs across the full audit pipeline can alleviate these bottlenecks by supporting human-in-the-loop workflows, enhancing scalability and enabling more holistic accessibility.

To address them, an integrated strategy is proposed to explore the competence of \underline{M}LLMs \underline{a}s \underline{C}opilots (MaC).

\textbf{Assistant}: \textit{Automating Labor-Intensive Tasks through Multimodal Reasoning}.
We explore two key use cases:
\ding{182}\textit{Individuality-based Page Sampling}:
The WCAG-EM relies on manually identifying structured sample pages based on individual factors like functional role and structural position.
By combining web crawling with MLLMs’ multimodal reasoning, we automate this process, enabling informed sampling that \textit{captures accessibility-critical Individuality beyond what is achievable through collective data-driven methods} like \grasp.
\ding{183}\textit{Pre-audit Element Localization}:
MLLMs can be used to preprocess pages, automatically identifying and labeling candidate elements for manual review.
This transforms manual auditing into a more efficient process of \textit{test without search}, where human experts validate elements without needing to navigate the page exhaustively.

\textbf{Auditor}: \textit{Identifying Underrepresented Accessibility Barriers}.
Recent studies highlight an overemphasis on visual and auditory disabilities in accessibility guidelines and tools, often \textit{overlooking cognitive  and situational disabilities that resist rule-based detection}~\cite{duo}.
MLLMs, with their contextual and inferential capabilities, offer a promising alternative.
We focus on WCAG 2.2 success criteria 3.3.8 and 3.3.9, which address accessible authentication and require reasoning about cognitive demands in user verification. 
MLLMs can help identify such mechanisms and assess potential barriers by interpreting page semantics at a higher level of abstraction than existing tools.

\textbf{Consultant}: \textit{Providing Informed Remediation Suggestions}.
The consultant role envisions MLLMs as intelligent agents for recommending fixes to accessibility issues, a direction supported by prior work on generating image descriptions, improving HTML semantics, and correcting ARIA attributes.
While promising, we focus on the assistant and auditor roles to address foundational scalability challenges, identifying the consultant role as a key avenue for future research in large-scale remediation.

In addition, we contend that \textit{\textbf{a major barrier to evaluating the effectiveness of MLLMs as Copilots is the lack of comprehensive, task-specific datasets tailored to accessibility}}.

\subsection{AWA: Datasets of AI for Web Accessibility}

To overcome this, we propose four novel datasets designed to advance the application of \underline{A}I for \underline{W}eb \underline{A}ccessibility (\data) in varirous tasks as shown in \tableautorefname~\ref{tab:tasks}.

\textbf{Triple-representativeness Page Sampling (TPS)}.
We have developed a dataset consisting of 495 publicly accessible websites, categorized into 117 distinct classes (please refer to Appendix).
These websites were crawled using an automated web crawler, ensuring that no private or sensitive data was involved.
On average, each website contains 196 webpages (ranging from a minimum of 104 to a maximum of 200), totaling 97,246 pages.
The dataset includes the following data for each page:
(1) the page’s DOM,
(2) a screenshot of the page,
(3) auto-check results covering 131 rules, where each rule corresponds to the number of violations detected (using Axe-core~\cite{axe}), and
(4) an adjacency matrix representing the website’s overall linkage graph. The inclusion of this novel adjacency matrix allows for more granular web accessibility sampling.

\textbf{Accessibility-relevant Page Recognition (APR)}.
To evaluate the ability of MLLMs Assistants in individuality-based page sampling, we constructed a manually annotated dataset consisting of 968 pages from five websites of different classes (entertainment, job search, e-commerce, government \& organizations, and social media), covering \textit{four category labels defined by WCAG-EM} for a structured sample:
(1) Common Web Pages and States,
(2) Relevant Web Pages and States,
(3) Pages of Essential Functionality, and
(4) Pages of Web Technologies Relied Upon for Conformance.
This dataset contains 951 human labels of four types with an equal distribution of positive and negative labels.
Fifty cases are selected as a few-shot fine-tuning set.

\textbf{CAPTCHA of Cognitive Tests (CCT)}.
Completely Automated Public Turing test to tell Computers and Humans Apart (CAPTCHA) is a widely utilized mechanism for distinguishing human users from automated bots, typically as part of login verification processes.
\textit{It presents various cognitive tasks designed to challenge and differentiate human recognition abilities from those of machines, making it an ideal test scenario for evaluating cognitive accessibility.}
In this study, we have collected a dataset of 1,985 CAPTCHA images from the internet, spanning 17 distinct categories of authentication requirements (see Appendix).
Among these categories, three meet the criteria outlined in WCAG 2.2 Success Criteria 3.3.9 for cognitive disabilities.
We also provide a 50\% train split drawn from each class for fine-tuning.

\textbf{Complete Process Extraction (CPE)}.
To assess MLLMs in Pre-audit Element Localization, we annotated 1,199 pages from the APR dataset, marking pages that contain \textit{five key components relevant to complete processes or accessibility defined by WCAG-EM}:
(1) search bar,
(2) select/filter panel,
(3) input form,
(4) CAPTCHA, and
(5) contact information.
598 positive labels and 601 negative labels are annotated.
Given the low occurrence frequency of some elements, we constructed 50 representative cases for few-shot fine-tuning, with an equal split between positive and negative examples.

\section{Experiments}

\subsection{Baselines and Experimental Settings}
\textbf{Baselines and Experimental Settings}:
For the \textit{collective page sampling}, we utilize the TPS dataset and compare \grasp with \textit{five statistical representations} proposed by a recent study on Web Structure Derived Clustering for Accessibility Page Sampling (SDC)\cite{ps} and its dimension reduction variants with t-SNE~\cite{tsne}.
We also assess two variants of \grasp: one leveraging GCN\cite{gcn} tailored for homophilic graphs, where nodes with similar labels tend to be interconnected, and another utilizing IGNN~\cite{ignn} for heterophilic graphs, where nodes of differing labels are more likely to be adjacent.
For the \textit{MaC}, we adopt the APR, CCT, and CPE datasets, with various models including GPT-4o (200B), GPT-4o-mini (8B), Qwen2.5-VL-72B (Qwen2.5), Intern2-VL-8B~\cite{inter}, and MiniCPM-V 8B (CPM)~\cite{cpm}.
See more details in Appendix.
\textbf{Metrics}:
SDC evaluates page sampling using the mean internal cosine similarity  of clusters, which reflects cluster cohesiveness but overlooks the quality of the \textbf{sampled} results.
We propose two enhanced metrics:
(i) \textbf{The mean inter-cluster cosine similarity $S_\text{sampled}$ of sampled nodes} in the layout and textual embedding spaces derived from BERT and ViT.
\textit{A lower value indicates greater diversity in sample nodes, suggesting more distinct representativeness}.
(ii) \textbf{the difference $D_\text{intra-inter}$ between the mean intra-cluster cosine similarity of all nodes and the mean inter-cluster cosine similarity of sampled nodes}.
\textit{A larger difference indicates that not only the clusters are internally cohesive but also sample nodes are distinct from one another.}
(2) For MaC tasks, we use accuracy for recognition and extraction on APR and CPE, and use precision, recall and macro F1-score for classification on CCT.

\subsection{Performance Analysis}
\begin{table}[t]
\small
\centering
    \setlength{\tabcolsep}{1mm}
    \begin{tabular}{lcccc}
    \toprule[1pt]
    \multirow{2}{*}{\textbf{Method}}& \multicolumn{2}{c}{\textbf{Layout Space} } & \multicolumn{2}{c}{ \textbf{Textual Space} } \\
     & $S_\text{sampled}$ & $D_\text{intra-inter}$ & $S_\text{sampled}$ & $D_\text{intra-inter}$ \\
    \midrule[0.8pt]
    \textbf{SDC\_content} & 56.66 & 9.96 & 89.29 & \underline{2.73} \\
    \textbf{+TSNE} & 55.14 & 6.46 & 88.32 & 1.66 \\
    \hline
    \textbf{SDC\_struc\_cont} & 55.61 & 11.53 & 89.59 & 1.93 \\
    \textbf{+TSNE} & 55.89 & 8.91 & 88.89 & 1.60 \\
    \hline
    \textbf{SDC\_structure} & 56.11 & 11.07 & 89.77 & 1.39 \\
    \textbf{+TSNE} & 55.93 & 10.07 & 89.16 & 1.51 \\
    \hline
    \textbf{SDC\_tags} & 54.18 & 10.76 & 88.76 & 2.12 \\
    \textbf{+TSNE} & 55.80 & 9.02 & 89.05 & 1.51 \\
    \hline
    \textbf{SDC\_tree} & 54.17 & 10.55 & 88.79 & 2.09 \\
    \textbf{+TSNE} & 55.86 & 8.81 & 88.85 & 1.63 \\
    \hline
    \textbf{GRASP\_GCN} & \underline{51.54} & \underline{13.05} & \underline{86.99} & 1.59 \\
    \textbf{GRASP\_IGNN} & \textbf{44.31} & \textbf{14.94} & \textbf{80.45} & \textbf{7.40} \\
    \bottomrule[1pt]
    \end{tabular}
\caption{Mean Performance across 495 Websites of \grasp on TPS.
The smallest $S_\text{sampled}$ and largest  $D_\text{intra-inter}$ are highlighted in \textbf{bold}, while the second are \underline{underlined}.
}
\label{tab:grasp}
\end{table}
\begin{table}[t]
\small
\centering
    \setlength{\tabcolsep}{1mm}
    \begin{tabular}{cccccc}
    \toprule[1pt]
     & \textbf{Category} & \textbf{GPT-4o} & \textbf{4o-mini} & \textbf{Qwen} & \textbf{CPM} \\
    \hline
\multirow{5}{*}{\rotatebox[origin=c]{90}{\textbf{Element}}} & Form & \textbf{90.48} & 88.10 & 35.44 & \underline{86.08} \\
& Contact & \underline{43.23 }& 16.13 & 38.67 & \textbf{60.93} \\
& Select/Filter & \underline{98.15} & 90.74 & \textbf{100} & 85.71 \\
& Search & \textbf{98.80} & 64.00 & \underline{89.39} & 79.18 \\
& CAPTCHA & \textbf{92.72} & 87.27 & \underline{92.00} & \underline{92.00} \\
\hline
\multirow{4}{*}{\rotatebox[origin=c]{90}{\textbf{\specialcell{Page}}}} & Com. & \underline{99.05} & \textbf{100} & 56.44 & 71.29 \\
& \specialcell{Ess.} & 82.09 & 82.09 & \textbf{90.63} & \underline{85.94} \\
& \specialcell{Rel.} & 22.95 & 7.10 & \textbf{84.92} & \underline{42.46} \\
& \specialcell{Tech.} & 77.50 & 5.83 & \textbf{93.91} & \underline{85.22} \\
    \hline
    \end{tabular}
\caption{Recall on APR and CPE Datasets.
Com., Ess., Rel. and Tech. denote \textit{Common Web Pages, Essential Functionality, Relevant Web Pages, and Web Technologies}, respectively.
}
\label{tab:ep1}
\end{table}

\begin{table}[t]
\small
\centering
    \setlength{\tabcolsep}{1mm}
    \begin{tabular}{cccccc}
    \toprule[1pt]
     & \textbf{Category} & \textbf{GPT-4o} & \textbf{4o-mini} & \textbf{Qwen} & \textbf{CPM} \\
    \hline
\multirow{5}{*}{\rotatebox[origin=c]{90}{\textbf{Element}}} & Form & \underline{68.64} & \textbf{86.98} & 32.70 & 46.54 \\
& Contact & \underline{54.64} & 47.68 & 53.26 & \textbf{56.51} \\
& Select/Filter & 78.70 & \textbf{89.81} & \underline{85.71} & 48.98 \\
& Search & \textbf{94.58} & 77.83 & \underline{85.99} & 65.70 \\
& CAPTCHA & 79.09 & \underline{93.64} & \textbf{95.00} & 87.00 \\
    \hline
\multirow{4}{*}{\rotatebox[origin=c]{90}{\textbf{\specialcell{Page}}}} & Com. & \underline{54.50} & \textbf{62.09} & 38.92 & 51.23 \\
& Ess. & \underline{63.43} & \textbf{79.10} & 52.76 & 51.97 \\
& Rel. & \underline{54.73} & 49.28 & \textbf{77.94} & 44.41 \\
& Tech. & \textbf{83.75} & 43.75 & \underline{47.60} & 43.48 \\
    \hline
    \end{tabular}
\caption{Precision on APR and CPE Datasets.
}
\label{tab:ep2}
\end{table}
\begin{table}[t]
\small
\centering
    \setlength{\tabcolsep}{1mm}
    \begin{tabular}{cccccc}
    \toprule[1pt]
     & \textbf{Category} & \textbf{GPT-4o} & \textbf{4o-mini} & \textbf{Qwen} & \textbf{CPM} \\
    \hline
\multirow{5}{*}{\rotatebox[origin=c]{90}{\textbf{Element}}} & Form & \underline{77.95} & \textbf{87.06} & 34.36 & 61.54 \\
& Contact & \underline{50.38} & 24.04 & 46.03 & \textbf{59.15} \\
& Select/Filter & \underline{87.60} & \textbf{89.91} & 87.50 & 62.69 \\
& Search & \textbf{98.01} & 77.29 & \underline{88.31} & 73.21 \\
& CAPTCHA & \textbf{95.33} & 93.20 & \underline{94.85} & 87.62 \\
    \hline
\multirow{4}{*}{\rotatebox[origin=c]{90}{\textbf{\specialcell{Page}}}} & Com. & \underline{68.65} & \textbf{72.41} & 47.90 & 59.26 \\
& Ess. & \underline{77.46} & \textbf{79.71} & 65.91 & 64.33 \\
& Rel. & 35.44 & 12.81 & \textbf{80.21} & \underline{44.57} \\
& Tech. & \textbf{87.32} & 9.40 & \underline{64.29} & 60.12 \\
    \bottomrule[1pt]
    \end{tabular}
\caption{F1 on APR and CPE Datasets.
}
\label{tab:ep3}
\end{table}

\textbf{\grasp Performance}.
The performance of \grasp is presented in \tableautorefname~\ref{tab:grasp}.
Several key observations can be made:
\textbf{First},\textit{ \grasp variants consistently yield more representative node samples}, evidenced by lower inter-cluster layout and textual semantic similarities $S_\text{sampled}$, as well as higher differences $D_\text{intra-inter}$.
\textbf{Second}, \textit{\grasp demonstrates superior performance when utilizing the heterophilic IGNN}, compared to the homophilic GCN.
This suggests that the linkage relationships within websites are more likely to exhibit diverse connections across distinct semantic clusters.
\textbf{Finally},\textit{the five representations of SDC show comparable performance to each other, with inclusion of t-SNE mostly improving representativeness}. 
GRASP\_IGNN demonstrates significantly better representativeness results across both textual and visual spaces, while SDC only performs relatively better in the textual space compared to GRASP\_GCN.

\textbf{\mac Assistant Performance}.
The performance of \mac in individuality-based page sampling and pre-audit element localization is presented in \tableautorefname~\ref{tab:ep1}, \ref{tab:ep2} and \ref{tab:ep3}.
\textbf{First}, \textit{the MLLM assistant demonstrates high accuracy}, exceeding 50\% for most high-level multimodal semantic understanding and recognition tasks, with several task types reaching over 90\%, indicating promising capabilities.
\textbf{Second}, \textit{larger MLLMs mostly outperform smaller MLLMs with the first or second ranks, although both large and small MLLMs exhibit varying preferences and strengths}.
GPT-4o excels in extracting smaller elements such as contact forms and search boxes, whereas GPT-4o-mini performs better with larger components like forms and CAPTCHAs.
This suggests that smaller MLLMs can also find effective use, \textit{highlighting the critical role of selecting an appropriate model or integrating multiple models to improve performance}.

\begin{table}[t]
\small
\centering
    \setlength{\tabcolsep}{0.5mm}
    \begin{tabular}{rccccccc}
    \toprule[1pt]
        \textbf{Model}  & \textbf{T.} & \textbf{Exist.}  & \textbf{P.} & \textbf{R.} & \textbf{F1} & \textbf{Vio.}  \\
    \midrule[0.8pt]
    \textbf{MiniCPM-sft\ \ 8B} &\checkmark& 85.64 & 15.00 & 10.88 & 11.72 & 38.20 \\ 
    \textbf{ Intern2-VL\ \ 8B} &\checkmark& \textbf{100} & \textbf{49.54} & \textbf{43.85} & \textbf{45.58} & \textbf{99.88}  \\
    \textbf{GPT-4o-mini\ \ 8B}  && 91.06 & 27.19 & 16.66 & 19.33 & 93.33  \\
    \midrule[0.8pt]
    \textbf{GPT-4o\ 200B} && 96.30 & 34.24 & 27.34 & 29.16 & 97.47 \\
    \textbf{Qwen2.5-VL\ 72B} && 90.75 & 39.79 & 32.47 & 34.55 & 84.96  \\ 
    \bottomrule[1pt]
\end{tabular}
\caption{Results on Dataset CCT.
\textit{T.} is short for training, while \textit{Exist.} means recall of existence of CAPTCHAs.
\textit{P., R., F1} are precision, recall and macro F1-score, respectively.
\textit{Vio.} is the accuracy of violation judgement of cogniton test.
}
\label{tab:cap-mean}
\end{table}
\begin{table}[t]
\small
\centering
    \setlength{\tabcolsep}{1mm}
    \begin{tabular}{ccccccc}
    \toprule[1pt]
        \textbf{Category} & \textbf{Exist.}  & \textbf{P.} & \textbf{R.} &\textbf{ F1} & \textbf{Vio.}   \\ 
    \midrule[0.8pt]
\textbf{1} & 100 & 100 & 100 & 100 & 100  \\ 
\textbf{2} & 100 & 50.00 & 25.00 & 33.33 & 100  \\ 
\textbf{3} & 100 & 50.00 & 43.75 & 46.67 & 100  \\ 
\textbf{4} & 100 & 100 & 100 & 100 & 100 \\ 
\textbf{5} & 100 & 20.00 & 19.27 & 19.63 & 99.09\\ 
\textbf{6} & 100 & 100 & 100 & 100 & 100  \\ 
\textbf{7} & 100 & 50.00 & 46.67 & 48.28 & 100  \\ 
\textbf{9} & 100 & 33.33 & 32.96 & 33.15 & 100  \\ 
\textbf{10} & 100 & 25.00 & 21.05 & 22.86 & 100  \\ 
\textbf{11} & 100 & 0 & 0 & 0 & 100  \\ 
\textbf{12} & 100 & 14.29 & 1.90 & 3.36 & 100  \\ 
\textbf{13} & 100 & 100 & 100 & 100 & 100  \\ 
\textbf{14} & 100 & 50.00 & 37.50 & 42.86 & 100  \\ 
\textbf{15} & 100 & 50.00 & 44.64 & 47.17 & 100  \\ 
\textbf{16} & 100 & 25.00 & 24.70 & 24.85 & 99.00 \\ 
\textbf{17} & 100 & 25.00 & 4.17 & 7.14 & 100  \\ 
    \toprule[1pt]
\textbf{Mean} & 100 & 49.54 & 43.85 & 45.58 & 99.88  \\ 
    \bottomrule[1pt]
\end{tabular}
\caption{Performance of Intern2-VL on CCT. 
\textit{Exist.} denotes the recall for CAPTCHA detection. 
\textit{Vio.} represents the accuracy of cognitive accessibility violation assessment.}
\label{tab:inter}
\end{table}
\textbf{\mac Auditor Performance}.
The results of the \mac on CAPTCHA cognition are presented in \tableautorefname~\ref{tab:cap-mean} with several key observations as follows.
(1) \textit{While the recognition of the existence of CAPTCHAs approaches 100\%, the classification results still leave considerable room for optimization.}
This may be partly attributed to the semantic similarities between the types.
However, even when CAPTCHAs are similar, the distinct cognitive tests and operational requirements can introduce different barriers.
For example, there are several initial recognition tasks, each followed by different subsequent tasks such as matching, segmentation, or recognition.
These variations may introduce different cognitive challenges.
(2) Nevertheless, \textit{MLLMs demonstrate high accuracy in determining whether CAPTCHAs might impede users with cognitive impairments.}
This suggests their reasoning capabilities regarding functionality and barriers remain strong, compensating for classification shortcomings.

More case studies are in the Appendix, e.g., limitations of small MLLMs in structured response and fine-tuned small MLLMs outperforming large untrained ones.

\section{Conclusion}
In response to the scalability challenges in web accessibility audits, we present a full-lifecycle WAA framework that operationalizes WCAG-EM through integration of Automation, AI, and Auditor (AAA).
Our contributions address critical resource bottlenecks in both sampling and evaluation, including GRASP for representative multimodal page sampling, MAC strategies for MLLM as Copilots for scalable WAA, and a suite of benchmark datasets.
They advance the state of scalable, WCAG-EM-aligned accessibility auditing and lay the groundwork for more scalable WAA practices.

\section*{Acknowledgments}
This work was supported by the National Natural Science Foundation of China (Grant No.62372408).

\bibliography{aaai2026}

\clearpage
\section*{Appendix}
\subsection*{Datasets, Experimental Settings and Prompts}

\subsubsection{Datasets Information}
\label{sec:dataset}

\begin{table*}[t]
    \centering
    \caption{Website Categories of TPS Dataset}
    \label{tab:categories}
   \resizebox{0.7\textwidth}{!}{
         \begin{tabular}{c|c}
        \toprule[1pt]
           \specialcell{(1)	Consumer Electronics \& Digital Devices\\(2)	Government Functions\\(3)	Advertising \& Marketing\\(4)	Government Portals\\(5)	Finance \& Economics\\(6)	Pharmaceuticals \& Pharmacy\\(7)	Hotels \& Hospitality\\(8)	Wedding \& Photography Studios\\(9)	Business Services\\(10)	Hardware \& Electrical Engineering\\(11)	E-Commerce Services\\(12)	Military \& National Defense\\(13)	Daily Chemicals \& Household Products\\(14)	Astronomy \& History\\(15)	Online Education\\(16)	E-Commerce Platforms\\(17)	Foreign Language Resources\\(18)	Personalized Messaging \& Avatars\\(19)	Food \& Culinary Services\\(20)	Search Engines\\(21)	Apparel \& Accessories\\(22)	Franchise \& Investment\\(23)	Humor \& Jokes\\(24)	Power \& Water Utilities\\(25)	Encyclopedias \& Dictionaries\\(26)	Automotive Platforms\\(27)	Banking \& Insurance\\(28)	Advertising Networks\\(29)	Libraries \& Exhibitions\\(30)	Logistics \& Transportation\\(31)	Software Downloads\\(32)	Broadcasting \& Telecommunications\\(33)	Collectibles \& Hobbies\\(34)	Household Items \& Accessories\\(35)	Food \& Beverage\\(36)	Driving Schools \& Training\\(37)	Real Estate Platforms\\(38)	Cybersecurity\\(39)	Gaming Platforms\\(40)	Web Directories\\(41)	Training \& Certification Institutions\\(42)	Software Applications \& Utilities\\(43)	Health \& Wellness\\(44)	Group Buying Platforms\\(45)	Chat \& Social Networking\\(46)	Web Portals\\(47)	Plumbing \& Security Systems\\(48)	Travel Agencies\\(49)	Domestic Services\\(50)	Cashback \& Price Comparison\\(51)	General Lookup Tools\\(52)	Design Resources\\(53)	Webmaster Tools \& Resources\\(54)	Auto Parts \& Accessories\\(55)	Entrepreneurship \& Investment\\(56)	Beauty \& Cosmetic Surgery\\(57)	Textiles \& Leather Products\\(58)	Mobile Devices \& Gadgets\\(59)	Technology \& Programming\\}  
           &   \specialcell{(60)	Hospitals \& Clinics\\(61)	Radio \& Television\\(62)	Primary \& Secondary Education\\(63)	IT News \& Information\\(64)	General Discussion Forums\\(65)	Online Lending Platforms\\(66)	Video \& Film Platforms\\(67)	Overseas Study \& Exchange\\(68)	Medical Devices \& Equipment\\(69)	Computer Hardware\\(70)	Astrology \& Horoscopes\\(71)	Ticketing \& Reservations\\(72)	Fiction \& Literature Platforms\\(73)	Lottery \& Sports Betting\\(74)	Commercial \& Department Stores\\(75)	Lifestyle Encyclopedias\\(76)	Classified Information\\(77)	Construction Materials\\(78)	Animation \& Comics Platforms\\(79)	Bicycles \& Motor Vehicles\\(113)	Data Analytics\\(80)	Regional Portals\\(81)	Maternal \& Infant Platforms\\(82)	Consumer Electronics \& Digital Devices\\(83)	Education \& Examination\\(84)	Higher Education Institutions\\(85)	Chemical \& Energy Industries\\(86)	Public Organizations\\(87)	Home Furnishing \& Building Materials\\(88)	Music Platforms\\(89)	Agriculture, Forestry,\\ Animal Husbandry \& Fisheries\\(90)	Social Sciences \& Humanities\\(91)	Automotive Manufacturers\\(92)	Celebrity \& Fan Communities\\(93)	Blog Platforms\\(94)	Email \& Communication Services\\(95)	Machinery \& Industrial Equipment\\(96)	Job Search \& Recruitment\\(97)	Law \& Regulations\\(98)	Public Institutions\\(99)	Multinational Corporations\\(100)	Domain \& Hosting Services\\(101)	Sports \& Athletics\\(102)	Electronic Components\\(103)	Shopping \& Product Reviews\\(104)	Social Networking Platforms\\(105)	Traffic \& Maps\\(106)	Packaging \& Printing\\(107)	Entertainment \& Fashion\\(108)	News \& Press Publications\\(109)	Petrochemical \& Energy\\(110)	Images \& Photography\\(111)	Educational Information\\(112)	Travel Platforms\\(114)	Securities \& Stock Platforms\\(115)	Pets \& Toys\\(116)	Ball Sports\\(117)	Electronic Payment Systems}\\
        \bottomrule[1pt]
        \end{tabular}
   }
\end{table*}
Categories of the TPS dataset are presented in \tableautorefname~\ref{tab:categories}, while 
17 CAPTCHA categories of the CCT datasets are documented in \tableautorefname~\ref{tab:captcha-app}.
All datasets and its detailed information are available at our open repository.

\subsubsection{Experimental Settings}
\grasp employs the open-source pre-trained models BERT and ViT  for the initial textual and visual embedding construction.
During training, the parameters of these two models are frozen, and only the parameters of the GNN and transformation matrices are updated.
For clustering, we apply the k-means algorithm and graph learning pipeline in HoLe uniformly across all baselines and \grasp, setting the number of clusters and sampled nodes to 20 and the iteration number of representativeness clustering and graph learning to 5 for consistency.
For the MLLMs, we perform fine-tuning on open-source small models with 8B parameters, while for GPT-4o-mini and larger models (Qwen2.5-VL-72B, GPT-4o), we conduct inference directly without training.

\subsubsection{Detialed Promopts}
\label{sec:prompts}
Detailed prompts for each datasets are presented in Prompt 1, Prompt 2 and Prompt 3.


\begin{table}[ht]
    \centering
    \label{tab:prompt3}
\rule{\columnwidth}{1pt}
\textbf{Prompt 3: CCT Dataset} 
\begin{lstlisting}[basicstyle=\tiny]
Please analyze the provided image information accurately and answer the following questions. Return the results in **JSON format**:

- **Question a**: Does the image contain a CAPTCHA? (Output 1 for yes, 0 for no)  
- **Question b**: If a CAPTCHA is present, which of the following 17 types does it belong to? (If not present, output 0)  
  1. Biometric verification  
  2. Click on a specific area  
  3. Slide a specific component  
  4. Drag a button to rotate an image to the correct angle  
  5. Drag an element to complete a puzzle  
  6. Swap elements to complete a puzzle  
  7. Object recognition (only visible and touchable objects)  
  8. Personal content recognition (user-provided content such as images, text, etc.)  
  9. Concept recognition (recognizing abstract concepts, text, graphics, etc.)  
  10. Matching elements such as graphics, text, or patterns provided by the website  
  11. Verification involving domain-specific conceptual knowledge  
  12. Composite verification involving multiple cognitive requirements  
  13. Draw along a specified path  
  14. Object segmentation  
  15. Mathematical operations  
  16. Cross-device verification, such as scanning a QR code or sending/receiving text via SMS, email, etc.  
  17. Other types of CAPTCHA not covered above, or confusing CAPTCHA types that cannot be clearly classified  
- **Question c**: Does the CAPTCHA fail to meet WCAG 2.2 AA standards and the cognitive function test requirements? (Output 1 if it fails, 0 if it meets the standard)  
- **Question d**: If the answer to question c is 1, briefly explain why it does not comply; otherwise, return an empty string.  

Ensure that the output is in standard JSON format, for example:
```json
{
  "a": 1,
  "b": 10,
  "c": 1,
  "d": "The CAPTCHA requires users to recognize specific objects or abstract concepts, which may pose challenges for users with cognitive disabilities and does not comply with WCAG 2.2.",
}
```
\end{lstlisting}
\rule{\columnwidth}{1pt}
\end{table}

\begin{table}[ht]
    \centering
    \label{tab:prompt1}
\rule{\columnwidth}{1pt}
\textbf{Prompt 1: APR and CPE Dataset} 
\begin{lstlisting}[basicstyle=\tiny]
1. Please analyze the provided webpage screenshot and determine whether the page **contains** the specified element.  
   - **Element Name**: Search Box  
   - **Element Description**: Check whether the page includes a search box element, such as an input field for searching or a search button.  
   - **Output**: `"Contains"` or `"Does Not Contain"`
2. Please analyze the provided webpage screenshot and determine whether the page **contains** the specified element.  
   - **Element Name**: Filters  
   - **Element Description**: Check whether the page includes filtering elements, such as filter buttons, filter criteria, etc.  
   - **Output**: `"Contains"` or `"Does Not Contain"`
3. Please analyze the provided webpage screenshot and determine whether the page **contains** the specified element.  
   - **Element Name**: Contact Information  
   - **Element Description**: Check whether the page includes ways to contact the site owner or staff, such as QR codes, phone numbers, email addresses, or physical addresses.  
   - **Output**: `"Contains"` or `"Does Not Contain"`
4. Please analyze the provided webpage screenshot and determine whether the page **contains** the specified element.  
   - **Element Name**: Form  
   - **Element Description**: Check whether the page includes any form elements, such as login, registration, or submission forms.  
   - **Output**: `"Contains"` or `"Does Not Contain"`
5. Please analyze the provided webpage screenshot and determine whether the page **contains** the specified element.  
   - **Element Name**: CAPTCHA  
   - **Element Description**: Check whether the page includes any CAPTCHA elements, such as slider puzzles, image-based CAPTCHAs, or SMS verification codes.  
   - **Output**: `"Contains"` or `"Does Not Contain"`
6. Please analyze the provided webpage screenshot and determine whether the page belongs to the specified page type.  
    - **Page Type**: Common Web Pages  

\end{lstlisting}
\rule{\columnwidth}{1pt}
\end{table}

\begin{table}[ht]
    \centering
    \label{tab:prompt2}
\rule{\columnwidth}{1pt}
\textbf{Prompt 2: APR and CPE Dataset} 
\begin{lstlisting}[basicstyle=\tiny]
   - **Type Description**: Identify whether the page is a common web page or a typical web application state. These pages are usually directly accessible from the homepage or navigated to via headers, navigation bars, or footers. Examples include the homepage, landing pages of submodules, login pages, and category pages.  
   - **Output**: `"Yes"` or `"No"`
7. Please analyze the provided webpage screenshot and determine whether the page belongs to the specified page type.  
   - **Page Type**: Essential Functionality  
   - **Type Description**: Identify whether the page supports the core functionality of the website. While some functions are obvious, others may require more exploration. The goal is to recognize key actions users can perform on the site. Examples include product purchasing pages in an online store, registration pages, video playback pages, or survey submission pages.  
   - **Output**: `"Yes"` or `"No"`
8. Please analyze the provided webpage screenshot and determine whether the page belongs to the specified page type.  
   - **Page Type**: Relevant Web Pages  
   - **Type Description**: Identify whether the page is particularly relevant to accessibility or people with disabilities. This includes pages about accessibility features, help and support, language or font settings, privacy policies, or contact details. These may or may not already be identified as part of the common pages.  
   - **Output**: `"Yes"` or `"No"`
9. Please analyze the provided webpage screenshot and determine whether the page belongs to the specified page type.  
   - **Page Type**: Web Technologies  
   - **Type Description**: Identify the key web technologies used on the page that affect accessibility. This includes fundamental technologies like HTML and CSS, assistive technologies like JavaScript and WAI-ARIA, or specialized technologies like SMIL, SVG, and PDF. Examples include pages using slider CAPTCHA, interactive maps, offline tools like Google Docs, or live data maps.  
   - **Output**: `"Yes"` or `"No"`
\end{lstlisting}
\rule{\columnwidth}{1pt}
\end{table}

\subsection*{Empirical Lessons from Case Studies}

\textbf{Limitations of Small MLLMs in Structured Response}. 
Small-scale MLLMs (e.g., 8B) often struggle to generate well-structured or instruction-compliant answers without fine-tuning.
For example, when prompted to provide binary outputs (e.g., a simple 'yes' or 'no') to facilitate downstream parsing, large models generally comply, whereas small models frequently embed the answer within lengthy and verbose explanations, making answer analysis difficult. 
In more complex scenarios, such as analyzing CAPTCHA images from the CCT dataset—where the model must determine the presence of a CAPTCHA, identify its type, and assess its WCAG compliance, we require responses to be formatted as a JSON object with three fields (see detailed prompts in Appendix).
Without fine-tuning, small models tend to ignore this instruction, returning unrelated or overly verbose explanations, or even refusing to answer.
\textit{However, after a brief fine-tuning phase, these models learn to adhere to the required structure.}
While minor formatting errors may persist, the responses become reliably parseable.
\textit{This suggests that for tasks demanding specific output structures, small MLLMs can be effectively adapted via light fine-tuning, mitigating one of their key usability challenges.}

\textbf{Fine-Tuned Small Models Can Outperform Larger Untrained Models}.
In tasks requiring complex multimodal reasoning, such as multi-stage interpretation and response generation over CAPTCHA inputs, \textit{small fine-tuned models demonstrated surprising capabilities}. 
Specifically, after being trained on 50\% of the CCT dataset, Intern2-VL with 8B parameters significantly outperformed other larger baselines, including GPT-4o (200B), achieving nealy 100\% in the reasoning of CAPTCHA existence and WCAG conformance within this specialized task.
\textit{This highlights the latent potential of smaller models to serve as domain-specific experts in accessibility, provided they are trained on high-quality, targeted datasets.}
The implications are twofold: first, it emphasizes the value of domain-adapted datasets in unlocking performance gains; and second, it points to a resource-efficient pathway for deploying MLLMs in accessibility applications.
Compared to their larger counterparts, small models are cheaper to deploy and easier to integrate into public service infrastructures, making them promising candidates for scalable, inclusive, and democratized accessibility solutions.

\subsection*{Addtional Results}

Various MLLM results on different CAPTCHA types are presented in \tableautorefname~\ref{tab:1}, \ref{tab:2}, \ref{tab:3}, \ref{tab:4} and \ref{tab:5} with corresponding type name of category ids in \tableautorefname~\ref{tab:captcha-app}.

\begin{table}[ht]
    \centering
    \small
    \setlength{\tabcolsep}{1mm}
        \begin{tabular}{clc}
    \toprule[1pt]
         \textbf{id} & \textbf{Type} & \textbf{P.}\\
    \midrule[0.8pt]
        1 & Biometric verification & \checkmark \\
        \hline
        2 & Click on a specific area & \\
        \hline
        3&Slide a specific component&\\
        \hline
        4&\specialcell{Drag a button to rotate images to the correct angle}&\\
        \hline
        5&Drag an element to complete a puzzle&\\
        \hline
        6&Swap elements to complete a puzzle&\\
        \hline
        7&\specialcell{Object recognition \\(only visible and touchable objects)}&\\
        \hline
        8&\specialcell{Personal content recognition of user-provided\\  content such as images and text}& \checkmark\\
        \hline
        9&\specialcell{Concept recognition of abstract \\ concepts, text, graphics}&\\
        \hline
        10&\specialcell{Matching elements such as graphics, text or \\ patterns, provided by the website}&\\
        \hline
        11&\specialcell{Verification involving domain-specific knowledge}&\\
        \hline
        12&\specialcell{Composite verification involving \\ multiple cognitive requirements}&\\
        \hline
        13&Draw along a specified path&\\
        \hline
        14&Object segmentation&\\
        \hline
        15&Mathematical operations&\\
        \hline
        16&\specialcell{Cross-device verification, inlcuding scanning \\a QR code, sending or receiving text via SMS and\\ email that allows copy and paste.}& \checkmark\\
        \hline
        17&\specialcell{Other types of CAPTCHA not covered above, or  \\confusing  CAPTCHA types that cannot be\\  clearly classified} &\\
    \bottomrule[1pt]
    \end{tabular}
\caption{CAPTCHA Categories in the CCT Dataset.
\textit{P.} indicates compliance with WCAG 2.2 Success Criterion 3.3.9 (Accessible Authentication - Enhanced, Level AAA).
Notably, \textit{7 Object Recognition category} also satisfies Criterion 3.3.8 (Accessible Authentication - Minimum, Level AA).
}
    \label{tab:captcha-app}
\end{table}

\begin{table}[ht]
\centering
\small
    \begin{tabular}{ccccccc}
    \toprule[1pt]
        Category & Exist.  & P. & R. & F1 & Vio.   \\
    \midrule[0.8pt]
\textbf{1} & 100 & 100 & 100 & 100 & 93.33  \\
\textbf{2} & 100 & 50.00 & 33.33 & 40.00 & 100  \\
\textbf{3} & 100 & 25.00 & 15.62 & 19.23 & 100  \\
\textbf{4} & 100 & 50.00 & 46.15 & 48.00 & 100  \\
\textbf{5} & 100 & 25.00 & 24.09 & 24.53 & 100  \\
\textbf{6} & 100 & 100 & 100 & 100 & 100  \\
\textbf{7} & 100 & 33.33 & 6.90 & 11.43 & 100  \\
\textbf{9} & 100 & 20.00 & 16.95 & 18.35 & 99.81  \\
\textbf{10} & 100 & 33.33 & 2.63 & 4.88 & 100  \\
\textbf{11} & 100 & 0 & 0 & 0 & 100  \\
\textbf{12} & 100 & 0 & 0 & 0 & 100  \\
\textbf{13} & 100 & 50.00 & 40.00 & 44.44 & 100  \\
\textbf{14} & 100 & 0 & 0 & 0 & 100  \\
\textbf{15} & 100 & 50.00 & 45.61 & 47.71 & 87.72  \\
\textbf{16} & 59.00 & 11.11 & 6.20 & 7.96 & 96.80  \\
\textbf{17} & 81.82 & 0 & 0 & 0 & 81.82  \\
    \toprule[1pt]
\textbf{Mean} & 96.30 & 34.24 & 27.34 & 29.16 & 97.47  \\
    \bottomrule[1pt]
\end{tabular}
\caption{GPT4o on CCT.
\textit{T.} is training, while \textit{Exist.} means recall of existence of CAPTCHAs.
\textit{P., R., F1} are \textit{precision, recall and macro F1-score}, respectively.
\textit{Vio.} denotes accuracy of violation judgment of cognition accessibility.
}
\label{tab:1}
\end{table}

\begin{table}[ht]
\centering
\small
    \begin{tabular}{ccccccc}
    \toprule[1pt]
        Category & Exist.  & P. & R. & F1 & Vio.  \\
    \midrule[0.8pt]
\textbf{1} & 66.67 & 50.00 & 33.33 & 40.00 & 86.67  \\
\textbf{2} & 100 & 50.00 & 16.67 & 25.00 & 100 \\
\textbf{3} & 100 & 25.00 & 20.31 & 22.41 & 93.75  \\
\textbf{4} & 100 & 33.33 & 15.38 & 21.05 & 92.31  \\
\textbf{5} & 99.54 & 20.00 & 17.63 & 18.74 & 88.58  \\
\textbf{6} & 100 & 100 & 100 & 100 & 100  \\
\textbf{7} & 100 & 33.33 & 12.64 & 18.33 & 93.10  \\
\textbf{9} & 98.70 & 9.09 & 0.03 & 0.07 & 92.19  \\
\textbf{10} & 100 & 16.67 & 10.96 & 13.23 & 97.37  \\
\textbf{11} & 100 & 0 & 0 & 0 & 100  \\
\textbf{12} & 100 & 0 & 0 & 0 & 100  \\
\textbf{13} & 100 & 50.00 & 10.00 & 16.67 & 100  \\
\textbf{14} & 100 & 0 & 0 & 0 & 100  \\
\textbf{15} & 100 & 33.33 & 27.49 & 30.13 & 82.46  \\
\textbf{16} & 19.30 & 14.29 & 2.14 & 3.73 & 94.10 \\
\textbf{17} & 72.73 & 0 & 0 & 0 & 72.73  \\
    \toprule[1pt]
\textbf{Mean} & 91.06 & 27.19 & 16.66 & 19.33 & 93.33  \\
    \bottomrule[1pt]
\end{tabular}
\caption{GPT4o-mini on CCT.
}
\label{tab:2}
\end{table}

\begin{table}[ht]
\centering
\small

    \begin{tabular}{ccccccc}
    \toprule[1pt]
        Category & Exist.  & P. & R. & F1 & Vio.   \\ 
    \midrule[0.8pt]
\textbf{1} & 37.50 & 50.00 & 25.00 & 33.33 & 100  \\ 
\textbf{2} & 50.00 & 0 & 0 & 0 & 0  \\ 
\textbf{3} & 100 & 50.00 & 43.75 & 46.67 & 0  \\ 
\textbf{4} & 100 & 100 & 100 & 100 & 100  \\ 
\textbf{5} & 100 & 25.00 & 22.05 & 23.43 & 98.18  \\ 
\textbf{6} & 100 & 100 & 100 & 100 & 100  \\ 
\textbf{7} & 100 & 50.00 & 30.00 & 37.50 & 100  \\ 
\textbf{9} & 100 & 0 & 0 & 0 & 85.56  \\ 
\textbf{10} & 100 & 33.33 & 17.54 & 22.99 & 100 \\ 
\textbf{11} & 100 & 0 & 0 & 0 & 100  \\ 
\textbf{12} & 100 & 20.00 & 1.33 & 2.50 & 100  \\ 
\textbf{13} & 100 & 100 & 100 & 100 & 100  \\ 
\textbf{14} & 100 & 0 & 0 & 0 & 100  \\ 
\textbf{15} & 100 & 50.00 & 48.21 & 49.09 & 92.86  \\ 
\textbf{16} & 81.20 & 33.33 & 27.53 & 30.16 & 99.40  \\ 
\textbf{17} & 83.33 & 25.00 & 4.17 & 7.14 & 83.33  \\ 
    \toprule[1pt]
\textbf{Mean} & 90.75 & 39.79 & 32.47 & 34.55 & 84.96  \\ 
    \bottomrule[1pt]
\end{tabular}
\caption{Qwen2.5  on CCT.
}
\label{tab:3}
\end{table}

\begin{table}[ht]
\centering
\small

    \begin{tabular}{ccccccc}
    \toprule[1pt]
        Category & Exist.  & P. & R. & F1 & Vio.   \\ 
    \midrule[0.8pt]
\textbf{1} & 0 & 0 & 0 & 0 & 25.00  \\ 
\textbf{2} & 50.00 & 0 & 0 & 0 & 100 \\ 
\textbf{3} & 100 & 33.33 & 25.00 & 28.57 & 0  \\ 
\textbf{4} & 100 & 33.33 & 23.81 & 27.78 & 0  \\ 
\textbf{5} & 99.09 & 20.00 & 18.36 & 19.15 & 2.73 \\ 
\textbf{6} & 100 & 100 & 100 & 100 & 0  \\ 
\textbf{7} & 86.67 & 20.00 & 1.33 & 2.50 & 60.00  \\ 
\textbf{9} & 99.26 & 0 & 0 & 0 & 90.00  \\ 
\textbf{10} & 100 & 0 & 0 & 0 & 57.89  \\ 
\textbf{11} & 100 & 0 & 0 & 0 & 71.43  \\ 
\textbf{12} & 100 & 0 & 0 & 0 & 53.33  \\ 
\textbf{13} & 100 & 0 & 0 & 0 & 0  \\ 
\textbf{14} & 100 & 0 & 0 & 0 & 0  \\ 
\textbf{15} & 96.43 & 0 & 0 & 0 & 35.71  \\ 
\textbf{16} & 55.40 & 0 & 0 & 0 & 31.80  \\ 
\textbf{17} & 83.33 & 33.33 & 5.56 & 9.52 & 83.33 \\ 
    \toprule[1pt]
\textbf{Mean} & 85.64 & 15.00 & 10.88 & 11.72 & 38.20  \\ 
    \bottomrule[1pt]
\end{tabular}
\caption{MiniCPM on CCT.
}
\label{tab:4}
\end{table}

\begin{table}[ht]
\centering
\small

    \begin{tabular}{ccccccc}
    \toprule[1pt]
        Category & Exist.  & P. & R. & F1 & Vio.   \\ 
    \midrule[0.8pt]
\textbf{1} & 100 & 100 & 100 & 100 & 100  \\ 
\textbf{2} & 100 & 50.00 & 25.00 & 33.33 & 100  \\ 
\textbf{3} & 100 & 50.00 & 43.75 & 46.67 & 100  \\ 
\textbf{4} & 100 & 100 & 100 & 100 & 100 \\ 
\textbf{5} & 100 & 20.00 & 19.27 & 19.63 & 99.09\\ 
\textbf{6} & 100 & 100 & 100 & 100 & 100  \\ 
\textbf{7} & 100 & 50.00 & 46.67 & 48.28 & 100  \\ 
\textbf{9} & 100 & 33.33 & 32.96 & 33.15 & 100  \\ 
\textbf{10} & 100 & 25.00 & 21.05 & 22.86 & 100  \\ 
\textbf{11} & 100 & 0 & 0 & 0 & 100  \\ 
\textbf{12} & 100 & 14.29 & 1.90 & 3.36 & 100  \\ 
\textbf{13} & 100 & 100 & 100 & 100 & 100  \\ 
\textbf{14} & 100 & 50.00 & 37.50 & 42.86 & 100  \\ 
\textbf{15} & 100 & 50.00 & 44.64 & 47.17 & 100  \\ 
\textbf{16} & 100 & 25.00 & 24.70 & 24.85 & 99.00 \\ 
\textbf{17} & 100 & 25.00 & 4.17 & 7.14 & 100  \\ 
    \toprule[1pt]
\textbf{Mean} & 100 & 49.54 & 43.85 & 45.58 & 99.88  \\ 
    \bottomrule[1pt]
\end{tabular}
\caption{Internvl2-sft on CCT.
}
\label{tab:5}
\end{table}

\subsection*{Broader social impact}

\textbf{Scalable, standardized audits for web accessibility}: Our AAA framework operationalizes the international standard WCAG-EM through a human–AI collaboration paradigm, enabling scalable accessibility audits that reduce reliance on labor-intensive manual processes.

\textbf{Advance AI-driven accessibility research}: The MaC and GRASP components provide novel methodologies, accelerating AI-driven research in web accessibility, supporting more inclusive digital experiences for users with disabilities, particularly in underserved regions or domains.

\textbf{Open datasets for community impact}: Our open, real-world datasets provide a foundation for transparent benchmarking, fostering community-wide progress in WAA.

\textbf{Informing next-generation standards}: As WCAG-EM is undergoing a potential update, our work offers timely and technically grounded insights that can shape future accessibility standards, amplifying societal benefits at scale.

\end{document}